\documentclass[letterpaper, 10 pt, conference]{ieeeconf-modified}
\IEEEoverridecommandlockouts                              % This command is only needed if
\usepackage{graphics} % for pdf, bitmapped graphics files
\usepackage{mathptmx} % assumes new font selection scheme installed
\usepackage{times} % assumes new font selection scheme installed
\usepackage{amsmath} % assumes amsmath package installed
\usepackage{amssymb}  % assumes amsmath package installed

\usepackage{mathtools}
\usepackage{soul}  % only for highlighting
\usepackage[table,xcdraw]{xcolor}

\usepackage[numbers]{natbib}
\usepackage{graphicx}
\usepackage{textcomp}
\usepackage{array}
\usepackage{multirow}
\usepackage{hyperref}
\usepackage[capitalize]{cleveref}

\crefname{equation}{}{}  % no eq. (2) but (2)
\newcommand{\Joints}{q}
\newcommand{\JointsPath}{Q}
\newcommand{\point}{{}^0p}

\newcommand{\Rot}{\operatorname{Rot}}
\newcommand{\Trans}{\operatorname{Trans}}

\newcommand{\Trobot}{F}
\newcommand{\T}[2]{{}^{#1}T_{#2}}
\newcommand{\Tcamerarobot}{\T{\mathrm{c}}{0}}
\newcommand{\Trightmarker}{p_{\mathrm{r}}}
\newcommand{\Tleftmarker}{p_{\mathrm{l}}}

\newcommand{\unitvec}[1]{{}^{0}e^{#1}}

\newcommand{\funForward}{f}
\newcommand{\funMeasurement}{h}

\newcommand{\funDHpar}{\rho}

\newcommand{\CALpar}{\Theta}
\newcommand{\CALparFrames}{c}

\newcommand{\DHpar}{\rho}
\newcommand{\DHparR}{\rho_0}
\newcommand{\DHd}{d}
\newcommand{\DHtheta}{\theta}
\newcommand{\DHr}{r}
\newcommand{\DHalpha}{\alpha}
\newcommand{\DHbeta}{\beta}

\newcommand{\CPpar}{\kappa} % compliance
\newcommand{\CPalpha}{\CPpar^{\DHalpha}}
\newcommand{\CPbeta}{\CPpar^{\DHbeta}}
\newcommand{\CPtheta}{\CPpar^{\DHtheta}}

\newcommand{\Mass}{m}
\newcommand{\MassPos}{w}
\newcommand{\Masspar}{\nu}
\newcommand{\torque}{\tau}
\newcommand{\gravity}{g}
\newcommand{\weighting}{\lambda}

\newcommand{\prediction}{y}
\newcommand{\prior}{\Lambda_\mathrm{p}}
\newcommand{\MarkerMeasurements}{y}

\newcommand{\cumsum}{\mkern-2mu .\mkern-2mu.\mkern-2mu. \mkern-8mu + \mkern-2mu}

\usepackage{fancyhdr}
\fancypagestyle{FirstPage}{
\lfoot{\fontsize{7}{7}\selectfont \begin{center}\copyright2021 IEEE. Personal use of this material is permitted. \\
Permission from IEEE must be obtained for all other uses, in any current or future media, including reprinting/republishing this material for advertising or promotional purposes, creating new collective works, for resale or redistribution to servers or lists, or reuse of any copyrighted component of this work in other works.\end{center}}
}

\title{\LARGE \bf
Calibration of an Elastic Humanoid Upper Body and Efficient Compensation for Motion Planning
}
\author{Johannes Tenhumberg and Berthold Bäuml% <-this % stops a space
\thanks{The authors are with the Institute of Robotics and Mechatronics, German Aerospace Center (DLR), Münchenerstr. 20, Wessling, 82234, Germany \newline
\tt\small{\{johannes.tenhumberg, berthold.baeuml\}@dlr.de}
}
\thanks{This work was partly funded by the Bavarian Ministry of Economic Affairs, Regional Development and Energy, within the projects SMiLE (LABAY97) and SMiLE2gether (LABAY102).}
}

\begin{document}

\maketitle
\thispagestyle{empty}
\pagestyle{empty}

%%%%%%%%%%%%%%%%%%%%%%%%%%%%%%%%%%%%%%%%%%%%%%%%%%%%%%%%%%%%%%%%%%%%%%%%%%%%%%%%

\begin{abstract}
High absolute accuracy is an essential prerequisite for a humanoid robot to autonomously and robustly perform manipulation tasks while avoiding obstacles.
We present for the first time a kinematic model for a humanoid upper body incorporating joint and transversal elasticities.
These elasticities lead to significant deformations due to the robot's own weight, and the resulting model is implicitly defined via a torque equilibrium.
We successfully calibrate this model for DLR's humanoid Agile Justin, including all Denavit-Hartenberg parameters and elasticities.
The calibration is formulated as a combined least-squares problem with priors and based on measurements of the end effector positions of both arms via an external tracking system.
The absolute position error is massively reduced from 21\,mm to 3.1\,mm on average in the whole workspace.
Using this complex and implicit kinematic model in motion planning is challenging.
We show that for optimization-based path planning, integrating the iterative solution of the implicit model into the optimization loop leads to an elegant and highly efficient solution.
For mildly elastic robots like Agile Justin, there is no performance impact, and even for a simulated highly flexible robot with 20 times higher elasticities, the runtime increases by only 30\%.
\end{abstract}

\section{Introduction}
\thispagestyle{FirstPage}
A prerequisite for a humanoid robot to autonomously and robustly perform tasks in the real world is an accurate kinematic model of itself.
For grasping, the robot needs to precisely position its tool center point (TCP) relative to objects, and when planning collision-free motions in self-acquired 3D models of the environment~\cite{Wagner2013} the extension of the whole body has to be predicted correctly.

Humanoid robots are complex mechatronical systems built from lightweight components, which often leads to a significant deviation from a straightforward, so-called, \emph{geometric} kinematics and make it necessary to model \emph{non-geometric} effects like elasticities.
For our advanced humanoid robot DLR Agile Justin~\cite{Bauml2014}, e.g., the deviation from the geometric kinematics leads to a mean error of 21\,mm and a worst-case error as large as 61\,mm in the whole workspace rendering robust autonomous action almost impossible.

In this paper, we present a kinematic model incorporating joint elasticities and transversal elasticities -- to our knowledge for the first time for a complete humanoid upper body.
We provide a clear and concise derivation of the kinematic model from the torque equilibrium, explicitly state the resulting exact implicit equation, and show how it can be solved iteratively.
We successfully calibrate all Denavit–Hartenberg (DH) parameters and elasticities of this model for the humanoid Agile Justin and discuss contributions of the components of the kinematic model to the error reduction.
Finally, we present our elegant approach to efficiently use the complex and implicit model in optimization-based motion planning with almost no performance impact.

\begin{figure}[t]
	\centering
	\includegraphics[width=0.6\linewidth]{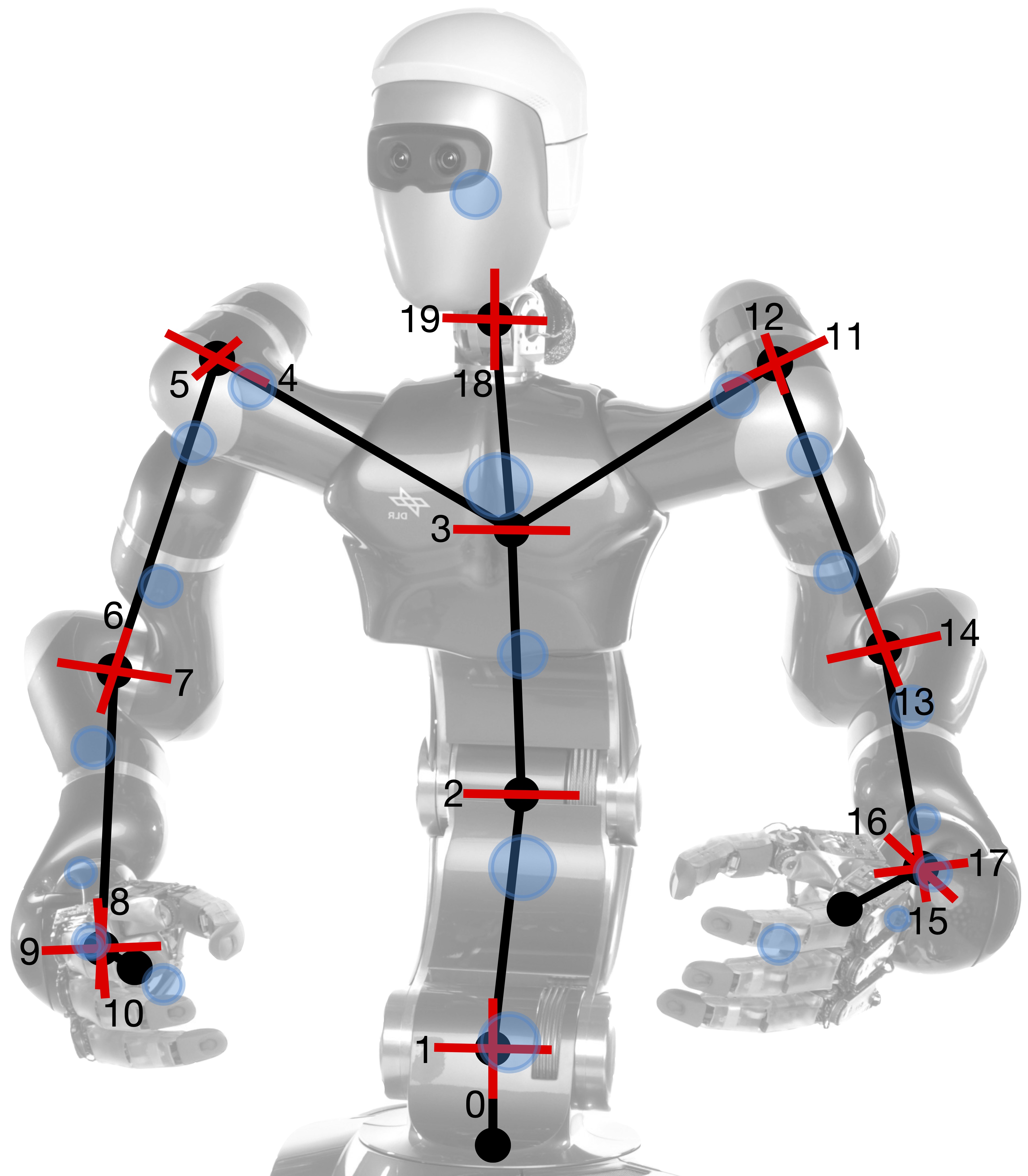}
	\caption{Kinematic tree structure of the humanoid Agile Justin, showing its 19 degrees of freedom (red) and the mass model used for calibration (blue).}
	\label{fig:justin_with_masses}
\end{figure}

\section{Related Work}
\citet{Journal1987} provide an early overview on robot calibration, where they describe four critical aspects:
the calibration model, the dataset, the identification of the model's parameters, and the compensation, i.e., how to use the model in a robotic task.
They also describe three different levels of calibration.
The first level is only finding the joint offsets.
The second is identifying all the robot's geometric parameters.
And the third level is the non-geometric calibration, including joint elasticities and gear backlash.

The first two calibration levels have been successfully applied to various robot arms~\cite{Pathre1990, Ginani2011, Park2011}.
In all these works, the focus is on algorithmic differences, what sets of DH parameters to use, and the comparison of the speed for calibration, but they ignore the speed of compensation.

Besides robotic arms, also more complex humanoid robots have been calibrated.
\citet{Maier2015} calibrate the joint offsets for the humanoid robot Nao and
\citet{Stepanova2019} all the DH parameters for the iCub robot.
They also investigate how to combine different internal measurement chains to get a full-body calibration.

Nevertheless, there are cases for which a purely geometric model of the robot is not sufficient.
A common source of non-geometric errors is the elasticities in the joints.
If the robot is equipped with joint torque sensors, it is convenient to use their measurements in the kinematic model as done in \citet{Klodmann2011} for the MIRO robot arm and by \citet{Besset2016} for an LWR arm.
But as the actual sensor readings are needed,  this approach is only feasible for control but not for motion planning.
Also, joint torque sensors can not measure any transversal effects.

A different approach for handling elasticities requires identifying a mass model of the robot and computing the torques at their static equilibrium for each pose.
\citet{Caenen1990} showed how to incorporate torques into the DH-formalism by adding torque-dependent offsets to the rotational DH parameters. However, they completely neglected to find the torque equilibrium.
Others~\cite{Khalil2002, Lee2013, Zhou2014b} included an iterative search to find the equilibrium, but they only dealt with the joint elasticities and did not include transversal elasticities.
Furthermore, they neglected the problem of efficient compensation of such an iterative model.

In the case of our elastic humanoid Agile Justin~\cite{Bauml2014},
an efficient compensation of the non-geometric model is crucial as we want to use it in an optimization-based path planner \cite{Wagner2013}.
In previous work, we already automatically calibrated the multi-sensorial head~\cite{Carrillo2013, Birbach2015} and the IMUs in the head and the base~\cite{Birbach2014}.
Instead, in this paper, we provide an accurate calibration and efficient compensation for Agile Justin's full kinematic as needed for whole-body motion planning.

\section{Robot Model}
\subsection{Geometric Model}
The forward kinematics of a robot maps from the configuration space of generalized joint angles to the robots' physical pose in the cartesian workspace.
This function is central to robotic path planning.
E.g., grasping an object and checking for obstacle or self-collision all strongly depend on an accurate model of the robot's kinematics.
A widely used representation of this kinematic model is formulated with the DH parameters.
In this formulation, four values $\rho_i= [\DHd_i, \DHr_i, \DHalpha_i,\DHtheta_i]$ describe the connection between two consecutive frames of the robot:
\begin{align}
\T{i-1}{i} =
\Rot_{\mathrm{x}}(\DHalpha_{i}) \cdot
\Trans_{\mathrm{x}}(\DHr_{i}) \cdot
\Rot_{\mathrm{z}}(\DHtheta_i) \cdot
\Trans_{\mathrm{z}}(\DHd_i)
\end{align}
The joints $\Joints_i$ are treated as offsets to $\DHtheta_i$ or $\DHd_i$ in \cref{eq:DH5} depending on the type of joint.
This minimal representation with two translational and two rotational parameters is enough to describe an arbitrary robot.
However, a limit of this formulation shows up for parallel axes (see joints 1 to 3 in \cref{fig:justin_with_masses}).
In this case, it is impossible to represent a small variation at the next link with a small variation of the DH parameters.
As \cite{Caenen1990} showed, a solution is to use an additional parameter $\DHbeta$ representing a rotation around the $y$-axis, leading to modified DH parameters $\rho_i= [\DHd_i, \DHr_i, \DHalpha_i, \DHbeta_i, \DHtheta_i]$:
\begin{align}
\T{i-1}{i}  =
\Rot_{\mathrm{y}}(\DHbeta_{i}) \cdot
\Rot_{\mathrm{x}}(\DHalpha_{i}) \cdot
\Trans_{\mathrm{x}}(\DHr_{i}) \cdot \nonumber \\
\Rot_{\mathrm{z}}(\DHtheta_i) \cdot
\Trans_{\mathrm{z}}(\DHd_i)
\label{eq:DH5}
\end{align}

The frame of the TCP ($i=M$) relative to the robot's base ($i=0$) is calculated by appling the transformations in series:
\begin{align}
\T{0}{M}  = \T{0}{1} \cdot \T{1}{2} \cdot \dotsc \cdot \T{M-1}{M}
\label{eq:TTT}
\end{align}
For more complex robots with a kinematic tree structure with multiple TCPs $M_i$, an equation like \cref{eq:TTT} holds for each branch.
We understand the forward kinematics $f$ to map from the robot configuration $\Joints$ not only to the position of the end effector(s) but to all frames  of the robot.
\begin{align}
\Trobot = [\T{0}{1}, \T{0}{2}, \dotsc, \T{0}{N}]   = f(\Joints, \DHpar)
\end{align}

\subsection{Non-geometric Model}
However, only considering the geometric model falls short of describing the real robot.
This paper focuses on torques as the primary source of non-geometric effects.
In general, an acting torque will bend the robot and produce a slightly different position.
As the DH parameters can describe an arbitrary robot, it is convenient to integrate the non-geometric effects into this formalism.
\citet{Caenen1990} showed this idea by explicitly expressing the influence of torques onto the DH parameters.
The simplest possibility is to add a torque-dependent linear to the geometric DH parameters $\DHparR$:
\begin{align}
\DHpar = \funDHpar(\DHparR, \CPpar, \torque) = \DHparR + \CPpar  \; \torque
\end{align}
Note, that the effects of forces on the link lengths are neglected and the matrix
$\CPpar = [0, 0, \CPalpha, \CPbeta, \CPtheta]$ describes only compliance around the respective axes, corresponding to the rotational DH parameters $[\DHalpha, \DHbeta, \DHtheta]$ and the acting torques $\torque = [\torque^x, \torque^y, \torque^z]$.
As a consequence, also the forward kinematics now depends via the DH parameters $\DHpar$ on the torques $\torque$ and the elasticity parameters $\CPpar$:
\begin{align}
\Trobot = \funForward(\Joints, \DHpar(\DHparR, \CPpar, \torque))
\end{align}

In the regime of no external forces and reasonably slow, quasi-static motions, the torques originate only from the robot's weight and its specific distribution.
For a frame, the pair $\Masspar_j=[\Mass_j, \MassPos_j]$ describes the mass $\Mass_j$ and its position $\MassPos_j$ relative to this frame (i.e., ${}^0 w_j = \T{0}{j} w_j$). The torques produced by this mass due to the gravity vector $\gravity$ around the respective coordinate axes of another frame $i$ with origin $\point_i$ can be calculated~\cite{Caenen1990} by\footnote{
Those torques act on the rotational axes parameterized by $\DHalpha_i$, $\DHbeta_i$, and $\DHtheta_i$, which are described in \cref{eq:DH5}. Because they make up a frame by sequential stacking, they do not all share the same frame of reference.
}:
\begin{align}
\torque^x_{ij} &= ((\T{0}{j} \, \MassPos_j - \point_{i-1}) \times \Mass_j \gravity) \cdot \unitvec{x}_{i-1} \nonumber \\
\torque^y_{ij} &= ((\T{0}{j} \, \MassPos_j - \point_{i-1}) \times \Mass_j \gravity) \cdot \unitvec{y}_{i-1} \\
\torque^z_{ij} &= ((\T{0}{j} \, \MassPos_j - \point_i)     \times \Mass_j \gravity) \cdot \unitvec{z}_{i}   \nonumber
\end{align}

To calculate the full torque $\torque_i$ acting on a link $i$, the contributions from all masses that act on the respective link have to be summed up, i.e., the torques from all masses which are higher up in the robot's kinematic tree.

The torques $\torque = \torque(\Trobot, \Masspar)$ now depend on the weight distribution $\nu = [\nu_1, \nu_2, \ldots]$ of the robot and therefore on the frames $\Trobot$ which are depending on the DH parameters $\DHpar$:
\begin{align}
\DHpar = \funDHpar(\DHparR, \CPpar, \torque(\Trobot(\Joints, \DHpar), \Masspar)) \label{eq:implicit_DH}.
\end{align}
This implicit equation for the DH parameters defines the equilibrium between the torques due to gravity and the torques due to flexion of the robot.
We define the solution to this equation as the non-geometric DH parameters
\begin{align}
\DHpar^* = \funDHpar^*(\Joints, \DHparR, \CPpar, \Masspar).
\end{align}

One possibility to solve \cref{eq:implicit_DH} is by the following iteration:
\begin{align}
\DHpar_n  &= (1-\weighting)\DHpar_{n-1} \; + \; \weighting \funDHpar(\DHparR, \CPpar, \torque(\Trobot(\Joints, \DHpar_{n-1}), \Masspar)) \label{eq:loop_DH} \\
\DHpar_{\infty} &= \DHpar^*
\end{align}
Choosing an appropriate $\lambda \in [0, 1]$ for the weighted sum ensures the convergence of the iteration even for very soft (or strongly non-linear) robots\footnote{The convergence can be shown by interpreting the  DH parameters as generalized coordinates and the update rule as the discretized  integration over time of an damped dynamical system (the robot moving due to gravity).}.
A suitable choice for the start of the iteration are the geometric DH parameters $\DHparR$
which can be interpreted as a robot in zero gravity or a robot with infinite stiffness.
Finally, the non-geometric forward kinematics is then given by
\begin{align}
\Trobot^* = \funForward^*(\Joints, \DHparR, \CPpar, \Masspar) := \funForward(\Joints, \DHpar^*(\Joints, \DHparR, \CPpar, \Masspar)).
\end{align}

\section{Calibration}
\subsection{Measurement Model}
\begin{figure}[tb]
    \centering
	\includegraphics[width=0.8 \linewidth]{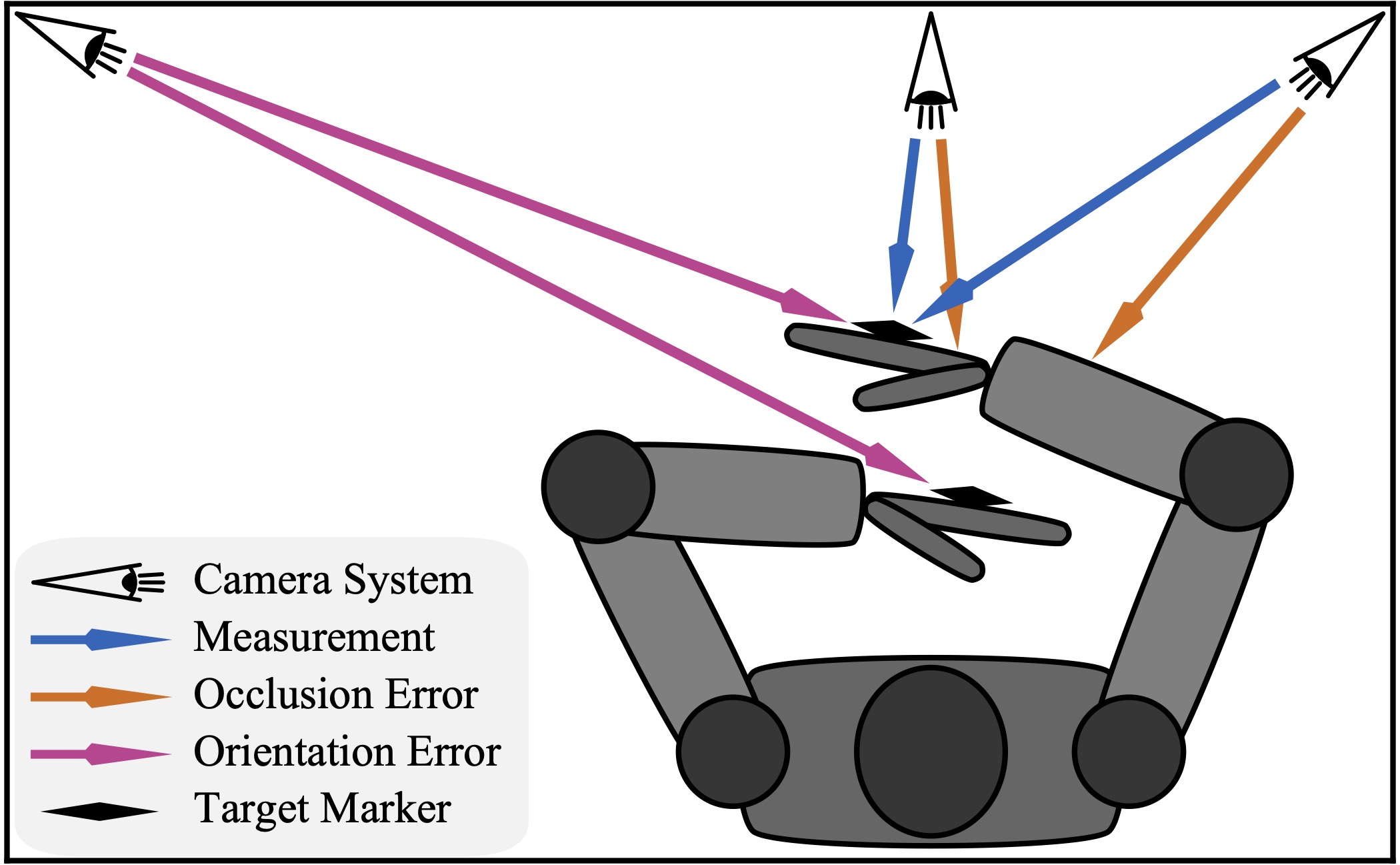}
	\caption{The measurement setup with cameras tracking markers on the end effectors; showcasing the problem of occlusion and suitable orientation of the marker for a given robot configuration.}
	\label{fig:calibration_sketch}
\end{figure}

We use an external camera system from Vicon, consisting of six 16Mpx cameras mounted on the ceiling and designed to track retro-reflective markers with high accuracy.
We fixated two such markers on the robot's hands, the last link of the respective kinematic chains (see \cref{fig:calibration_sketch}).
To use the markers' positions $\prediction$ for calibrating the forward kinematics $\funForward$, additional information is necessary.
First, we need the robot's base relative to the camera system's world frame $\Tcamerarobot$.
Second, the markers' position on the left and right end effector $\Trightmarker$ and $\Tleftmarker$ must be known.
It is not always possible to determine those frames beforehand; therefore, they become part of the calibration problem.
Our measurement function $\funMeasurement$ consists of the forward kinematics and the additional frames at the ends of the kinematic chain to close the measurement loop:
\begin{align}
y = \funMeasurement(\Joints, \CALpar) = \Tcamerarobot \cdot \funForward^*(\Joints, \DHparR, \CPpar, \Masspar)_{\mathrm{r}, \mathrm{l}} \cdot [\Trightmarker, \Tleftmarker]
\end{align}
The calibration parameters $\CALpar=[\DHparR, \CPpar, \Masspar, \CALparFrames]$ are a combination of the DH parameters $\DHpar$, the compliances $\CPpar$, the masses $\Masspar$, and the parameters $\CALparFrames= [p_{\mathrm{c}}, o_{\mathrm{c}} , \Trightmarker, \Tleftmarker]$ for the camera base-frame and marker positions to map the forward kinematics to the measurements.
The camera frame $\Tcamerarobot$ is here defined by its position $p_{\mathrm{c}}$ and its orientation $o_{\mathrm{c}}$.

\subsection{Identification}
The parameters $\CALpar$ of the measurement function $\funMeasurement$ can be identified using a dataset $D = \{(\Joints^{(n)}, \MarkerMeasurements^{(n)})\}_{n=1\ldots N}$ of pairs of the robot's configuration $\Joints^{(n)}$ and the corresponding marker positions $\MarkerMeasurements^{(n)}$.
We formulate the identification as a single combined least-squares problem based on all measurements and all markers to minimize the task space error.
To solve for the optimal parameters $\CALpar^*$, we use the maximum a posteriori (MAP) approach:
\begin{align*}
\min_{\CALpar} \left[\sum_n^N \frac{1}{\sigma_\mathrm{m}^2} |\prediction^{(n)} - h(\Joints^{(n)}, \CALpar)|^2 + (\CALpar - \CALpar_\mathrm{p})^T \prior^{-1} (\CALpar - \CALpar_\mathrm{p})\right]
\end{align*}
This approach uses a prior Gaussian distribution with mean $\Theta_{\mathrm p}$ and a diagonal covariance matrix $\prior = \operatorname{diag}{\sigma_{\mathrm{p}}^2}$, where the vector $\sigma_{\mathrm{p}}$ describes the uncertainty of the different calibration parameters. For the measurement noise, we use the usual assumption of a Gaussian distribution with zero mean and standard deviation $\sigma_\mathrm{m}$.
The prior acts as regularization and guarantees the existence of a minimum, even in the presence of redundancies in the measurement/kinematic model. Setting the prior is not critical, but it should be set conservatively to plausible values. 
Because of the highly nonlinear measurement function, it might be necessary to use multistart for the initial guess to find the global minimum.

\subsection{Configuration Selection and Efficient Sample Collection} \label{sec:sample_collection}
We selected the measurement poses randomly by sampling from the configuration space to ensure good overall accuracy and not only close to some standard configurations.
Nevertheless, the poses must be feasible for our experimental setup.
Besides avoiding self-collision while measuring, the markers imposed additional constraints, as they must be well visible to the cameras.
We used rejection sampling to ensure that at least four of the six cameras had a clear view of one marker.
We checked for occlusion with simple ray tracing, a straight line from each camera to the marker's position.
A sphere model of the robot, which we also use for collision checking, was used to test if the robot blocks any ray.
Combining all those constraints, only $1/10000$ configurations were feasible.

After determining a set of feasible configurations in simulation, we now need to measure those poses.
The robot has to move to each configuration so that the camera system can collect the corresponding marker positions.
One crucial concern regarding experiments with robots is always the time involved.
To perform short and collision-free paths from pose to pose, we use an optimization-based path planner; the same planner we want to make more accurate through calibration, but with extended safety margins for the nominal kinematic.

To reduce the time further, we ordered the randomly selected poses to minimize the distance between them.
By solving a traveling salesman problem on batches of 100 samples, we could reduce the time by a factor of two.
Now the average time to collect one sample is ten seconds.
From this, nine seconds fall on the robot performing the trajectory, and one second is added as a pause for the measurement.

\section{Compensation}
\begin{figure}[tb]
	\centering
	\includegraphics[width=0.80\linewidth]{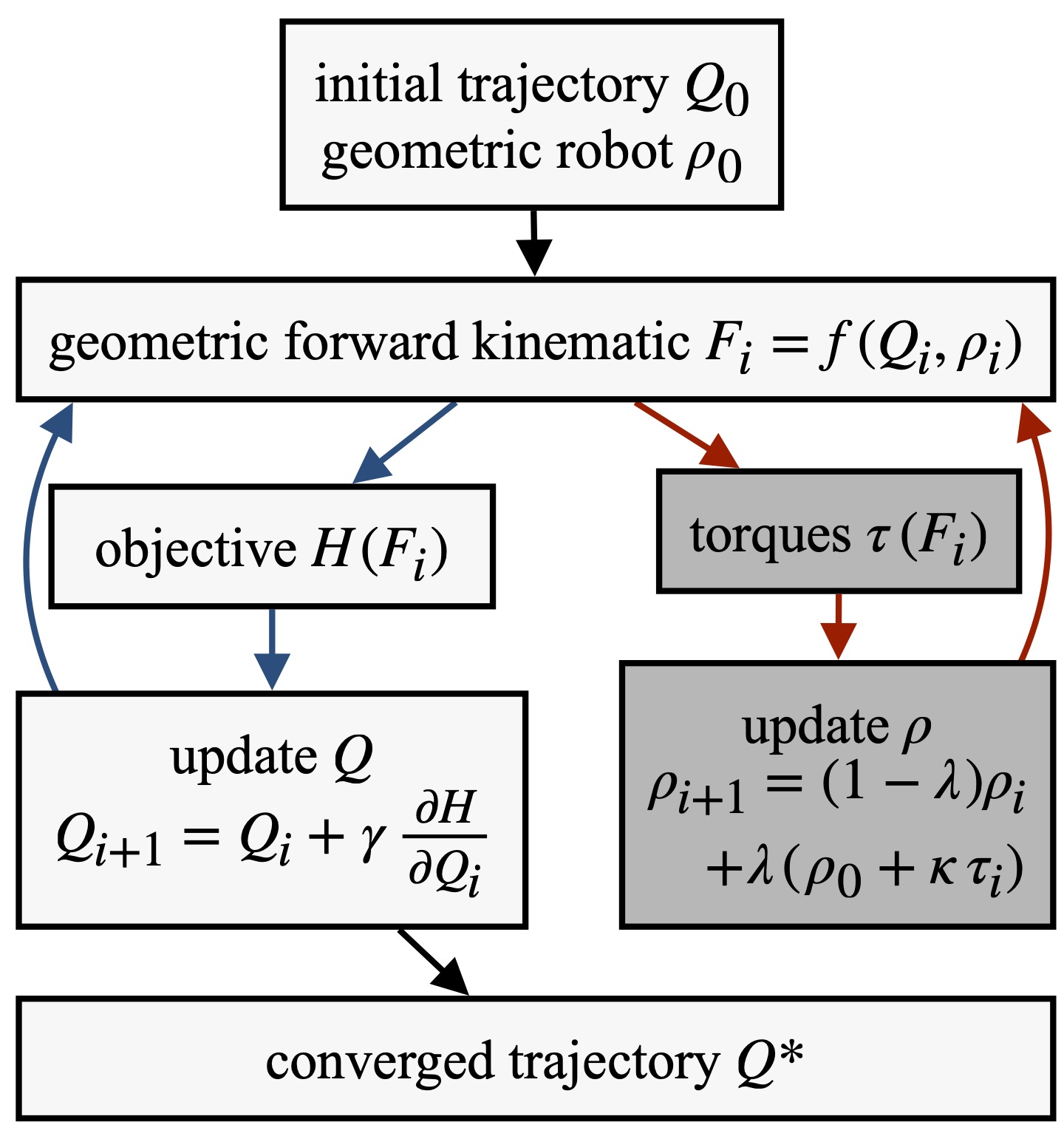}
	\caption{
	Flowchart of the optimization loop in light gray.
	Dark gray shows the additional loop for the static torque equilibrium.
	As the outer loop converges, no additional passes through the inner torque loop are necessary.
	}
	\label{fig:planning_and_torques_flowchart}
\end{figure}

The calibration goal is to find a set of parameters $\CALpar$ which describes the robot as accurately as possible.
Nevertheless, also the speed and ease of use of the calibration model, e.g. in motion planning, is crucial.
Incorporating the new set of DH parameters $\DHpar$ is straightforward and works without any changes or additional costs as they replace the old DH parameters.
The same holds for masses $\Mass$ and compliances $\CPpar$.
However, to determine the elastic effects, one must find the static equilibrium between acting torques and elasticities described in \cref{eq:implicit_DH}.
While it might be feasible for calibration (offline procedure) to use the iterative algorithm \cref{eq:loop_DH}, it is more prohibitive for compensation (online).
One should carefully evaluate this trade-off between accuracy and simplicity when choosing a calibration model for a robot.

Knowing that we will use the forward kinematics mostly in the framework of an optimization-based path planner has further implications.
Such a planner works on paths in configuration space $\JointsPath = [\Joints_{1}, \Joints_{2}, ..., \Joints_{n}]$ and performs iterations to get from an initial path $\JointsPath_{0}$ to a converged path $\JointsPath^*$.
For each step, the optimizer considers the objective function $H(\JointsPath_i)$ and updates the path using the gradient information.
The nested structure of the forward kinematics
\begin{align}
f^*(q) = f(q, \rho(f(q, \rho(f(...))))
\end{align}
leads to highly non-linear gradients, making the model more difficult to use.
As a solution to this, we separate the torque equilibrium search in a separate loop, as shown in \cref{fig:planning_and_torques_flowchart}.
After determining the acting torques accurately and updating the DH parameters accordingly, we assume these as constant for the gradient calculation $\partial \Trobot / \partial \Joints$.
This allows us to use the pure geometric forward kinematics, implicitly assuming $\partial \DHpar / \partial \Joints=0$.
Although we completely omit the torque iterations for the gradients, this approximation does not hinder convergence in our tests.

\begin{figure}[tp]
	\centering
	\includegraphics[width=\linewidth]{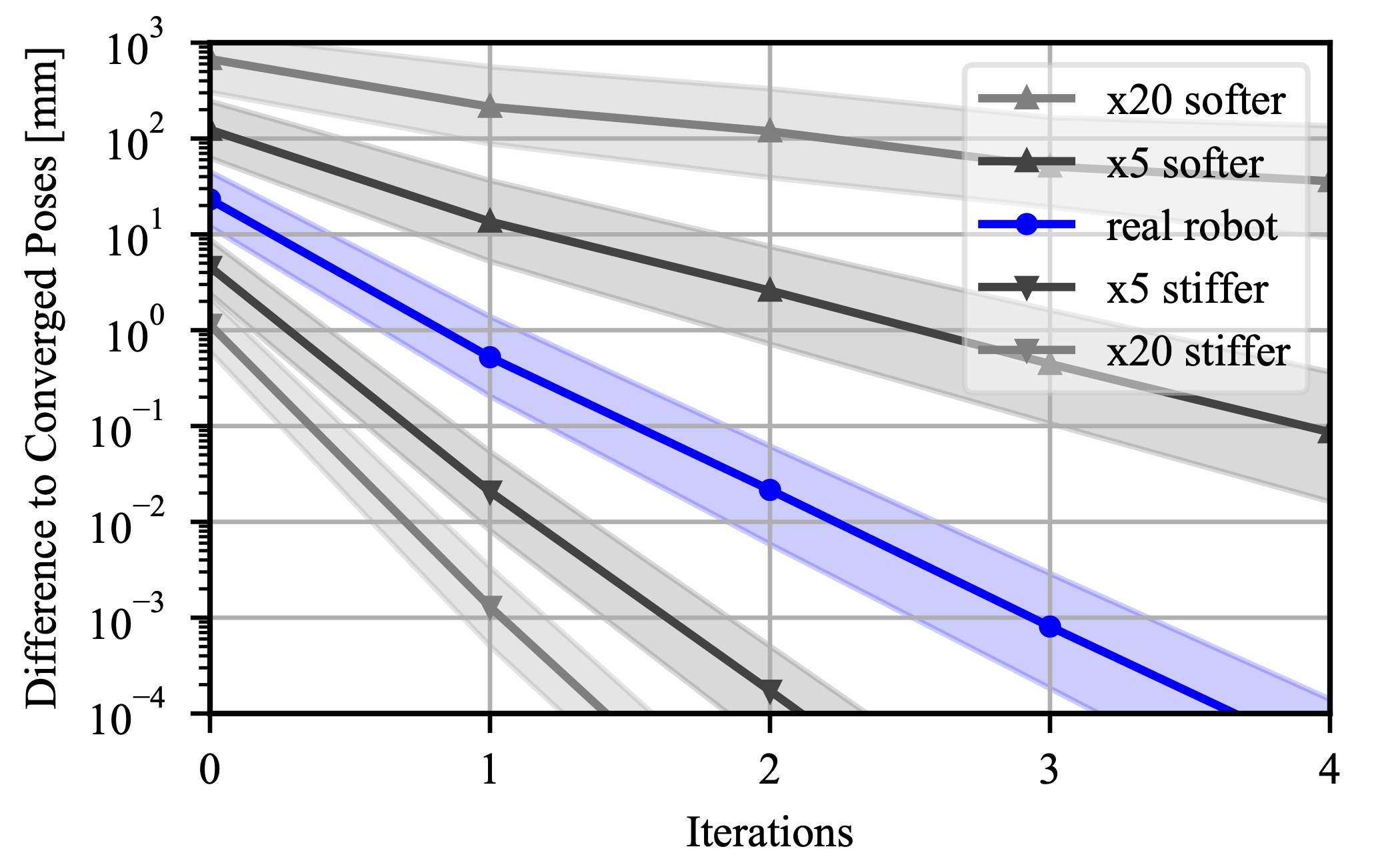}
	\caption{
	Simulated convergence towards the static torque equilibrium for robots of different stiffness.
    For iteration 0 the elastic effect is ignored.
	The bands indicate the standard deviation for 1000 different joint configurations.
	}
	\label{fig:torque_equilibrium}
\end{figure}

The second important aspect visualized in \cref{fig:planning_and_torques_flowchart} is the combination of the optimization loop and the torque equilibrium loop.
Even if the updates of the configurations are large at the beginning of the optimization, when the planner converges, the pose updates get small.
If those updates are significantly smaller than the offset produced by a torque update, it is unnecessary to search for a new static equilibrium iteratively.
In other words, it is sufficient to reuse the already computed frames, and calculate the acting torques and update the DH parameters only once, while nevertheless solving the forward kinematics $\funForward^*$ exactly.
The outer loop of the converging optimizer makes it unnecessary to perform inner iterations when searching the torque equilibrium.

\section{Experimental Evaluation}
\subsection{Calibration}
We collected $N_\mathrm{all} = 500$ samples with the procedure described in \cref{sec:sample_collection} and split them into a calibration set with $N=300$ and a test set with $N_\mathrm{test}=200$ samples. 
This data, as well as an overview of the nominal and the calibrated parameters, are provided online\footnote{\href{https://dlr-alr.github.io/dlr-elastic-calibration}{https://dlr-alr.github.io/dlr-elastic-calibration}}.

In what follows, we calibrate the parameters $\CALpar=[\DHparR, \CPpar, \CALparFrames]$, i.e., not $\Masspar$ (the mass $\Mass$ and the center of mass $\MassPos$) because they could be adopted from the CAD-files of the robot. 
In a test, we found that adding $\Masspar$ to the calibration parameters did not improve the residual error but significantly increased the runtime of the optimization algorithm.
The priors used for the following calibrations were chosen based on previous experiments with the robot, with uncertainty $\sigma_{\mathrm{p}}$ in lengths of 0.1\,m, angles of 0.2\,rad, and elasticities of 0.1\,rad/kNm.

In all experiments, we used multi start for the optimization with initial guesses $\CALpar_0$ randomly sampled from the prior distribution. 
All optimization runs converged to the same optimum, showing the stability of our calibration approach.

\begin{table}[t]
\caption{Calibration results: residual TCP errors [millimeters].}

\resizebox{\linewidth}{!}{%
\begin{tabular}{|c|c|cc|cc|}
\cline{2-6}
\multicolumn{1}{c|}{} & \textbf{Frames } & \multicolumn{2}{c|}{\textbf{DH parameter }} & \multicolumn{2}{c|}{\textbf{Compliance }} \\
\cline{2-6}
\multicolumn{1}{c|}{} & $\CALparFrames$  & \multicolumn{1}{c|}{\textbf{$\cumsum\DHtheta$ }} & \textbf{$\cumsum\DHd$, $\DHr$, $\DHalpha$, ($\beta$) } & \multicolumn{1}{c|}{\textbf{$\cumsum\CPtheta$ }} & \textbf{$\cumsum\CPalpha$, ($\CPbeta$) } \\
\hline
$\mu$    & 21.33 & 18.18 & 10.5  & 5.64  & 3.12 \\
\cline{1-1}
$\sigma$ & 9.71  & 7.8   & 5.74  & 2.57  & 1.71 \\
\cline{1-1}
$\max$   & 63.41 & 45.9  & 36.37 & 12.83 & 8.23 \\
\hline
\multicolumn{1}{c}{} & \multicolumn{1}{c}{\upbracefill} & \multicolumn{1}{c}{} & \multicolumn{1}{c}{} & \multicolumn{1}{c}{} & \multicolumn{1}{c}{\upbracefill} \\
\multicolumn{3}{c}{before calibration}  & \multicolumn{1}{c}{} & \multicolumn{2}{r}{full calibration}        \\
\end{tabular}
}

\label{tab:model_parameters__add_one}
\end{table}

\begin{figure}[t]
	\centering
	\includegraphics[width=\linewidth]{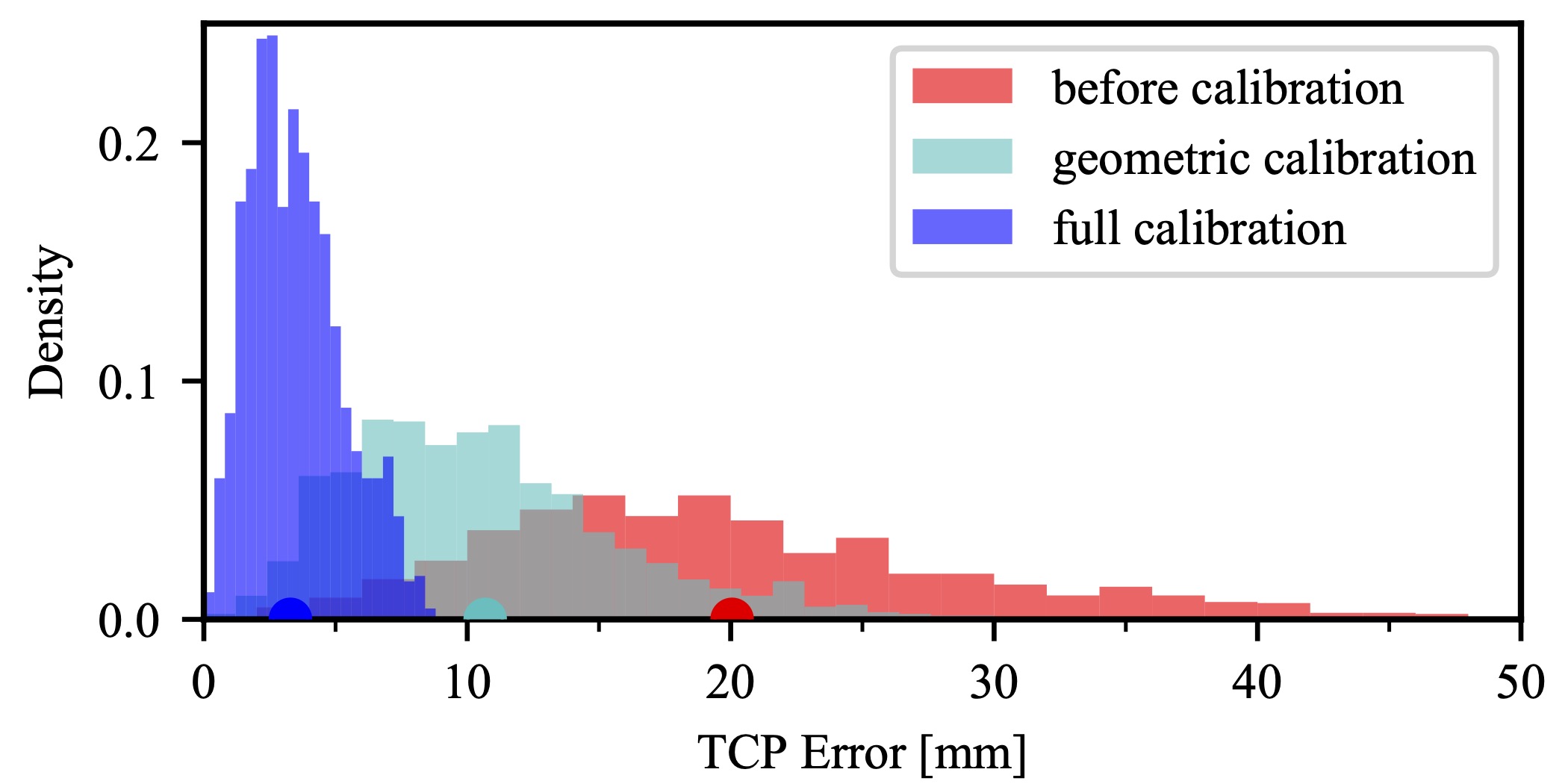}
	\caption{
	Histogram of the residual error at the end effectors before and after calibration, highlighting the need for non-geometric modeling of Justin.
	The circles on the $x$-axis mark the mean $\mu$ of the error (see also \cref{tab:model_parameters__add_one}).}
	\label{fig:error_hist}
\end{figure}

\begin{table}[t]
\caption{"Leave-one-out analysis": residual TCP errors [millimeters].}
\resizebox{\linewidth}{!}{%
\begin{tabular}{|c|c|cccccc|} 
\cline{2-8}
\multicolumn{1}{c|}{} & full & \multicolumn{1}{c|}{\textbf{$-\CPtheta$ }} & \multicolumn{1}{c|}{\textbf{$-\DHalpha$ }} & \multicolumn{1}{c|}{\textbf{$-\DHtheta$ }} & \multicolumn{1}{c|}{\textbf{$-\CPalpha$ }} & \multicolumn{1}{c|}{\textbf{$-\DHr$ }} & \textbf{$-\DHd$ } \\ 
\hline
$\mu$    & 3.12 & 9.09  & 6.48  & 6.35  & 4.85  & 4.33  & 3.46 \\ 
\cline{1-1}
$\sigma$ & 1.71 & 4.23  & 2.93  & 3.16  & 2.28  & 2.03  & 1.73 \\ 
\cline{1-1}
max      & 8.23 & 24.59 & 17.25 & 18.89 & 13.37 & 11.09 & 8.97 \\
\hline
\end{tabular}
}
\label{tab:model_parameters__leave_one_out}
\end{table}
\begin{figure}[t]
	\centering
	\includegraphics[width=\linewidth]{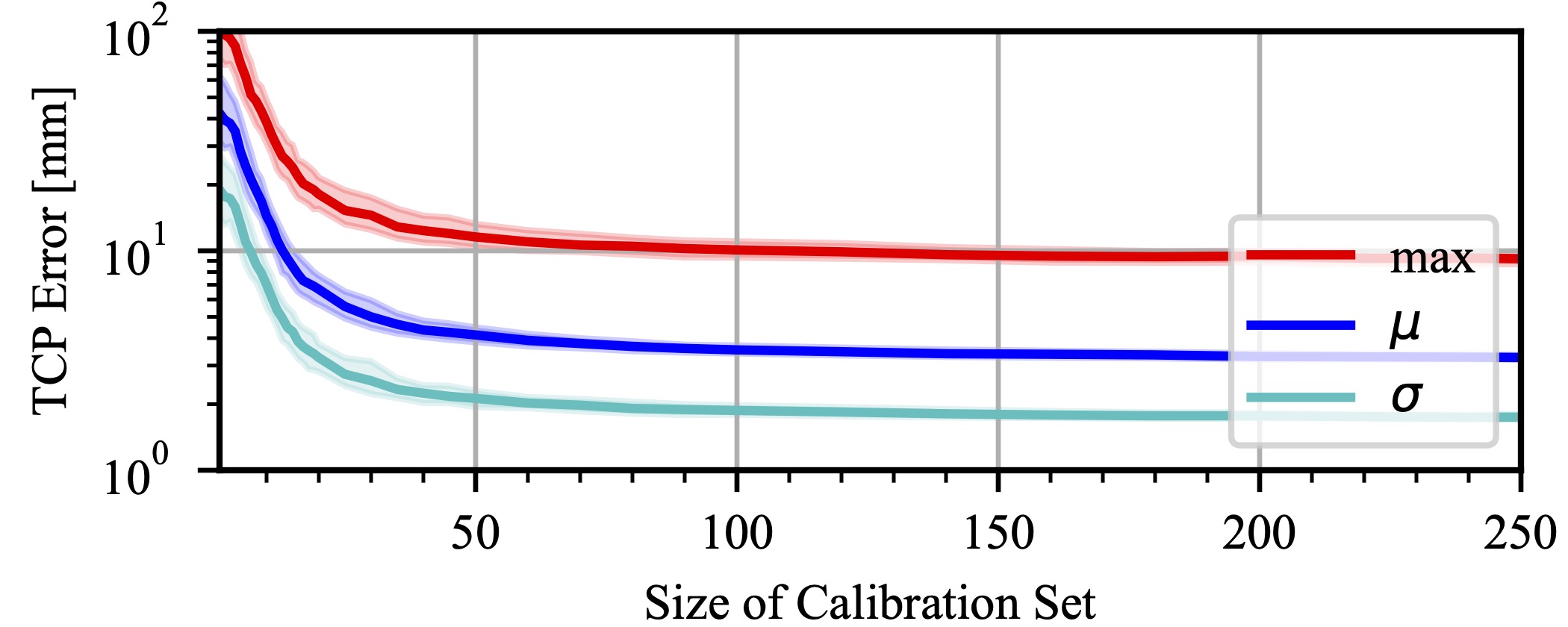}
	\caption{
	Test error over the calibration set size $N$. The bands show the standard deviation over 1000 different calibration sets of a given size.}
	\label{fig:size_of_calibration_set}
\end{figure}

As the first step, the additional frames parametrized by $\CALparFrames$ must be determined to close the measurement loop.
Only calibrating those frames gives the accuracy of the nominal kinematics before calibration, with a mean residual error of 20\,mm - averaged over the 200 test poses and both arms.
In the worst cases, the error was larger than 60\,mm.
After calibration, the mean residual error of the full model is reduced to 3.1\,mm and the maximum error to 8.2\,mm.
Those two cases are reported in \cref{tab:model_parameters__add_one} in the left- and rightmost columns and a detailed distribution of the errors is shown in \cref{fig:error_hist}.
Here the result of a pure geometric calibration without elasticities is also shown, which highlights the need for a non-geometric calibration model for this robot.

We conducted further experiments to evaluate the significance of the different model parameters.
\cref{tab:model_parameters__add_one} reports the residual error when making the model more expressive by adding more parameters step by step, starting with the uncalibrated model (leftmost column) up to the full model (rightmost column).
The order in which we added the different types of parameters (always for all joints at once) followed "standard practice" with joint offsets as first and transversal elasticities as the last addition.

That this "standard ordering" does not represent the actual influence of the parameters on the residual error can be seen in \cref{tab:model_parameters__leave_one_out}.
Here, starting from the full model (rightmost column), only a specific parameter type was left out one at a time.
From this, it can be seen how crucial joint and transversal elasticities are for modeling Justin's kinematics.
One reason for the joint elasticities being that prominent is that we not only model the mechanical elasticity of a joint but also the elasticity of the joint position controller~\footnote{
We use a relatively simple joint position controller as it is robust (e.g., it does not rely on the torque sensors, which are notoriously drifting and hard to maintain). However, it results in an additional joint-level elasticity.}.

\cref{fig:size_of_calibration_set} shows how the test error decreases when more samples $N$ are used for calibrating the full model.
Furthermore, one can see the influence of which robot configurations (joint angles) are in a set.
The light bands show the standard deviation when selecting 1000 different sets of a given size.
While there is still some minor improvement beyond $N=100$ samples,
the main effect of the calibration happens with as little as $N=50$ samples (those could be collected in less than ten minutes).
This information is especially beneficial for recalibrating the robot with a subset of the parameters in the future. A smaller calibration model corresponds to a smaller set necessary for calibration, making the effect of selecting an optimal set of configurations~\cite{Carrillo2013} more significant.

\subsection{Compensation}
\cref{fig:torque_equilibrium} visualizes the non-geometric effects in more detail by showing the convergence towards a static equilibrium for solving the kinematics via iteration according to ~\cref{eq:loop_DH}.
We show results for the real and virtual versions of Agile Justin with lower and higher elasticities.
For the real robot, the difference after just one iteration is already considerably below our calibration accuracy.
But if the robot is softer or the accuracy requirements are stricter, one needs more iterations to converge to the equilibrium configuration.

In addition, we tested the algorithm described in \cref{fig:planning_and_torques_flowchart} on these versions of Agile Justin.
While for the real or stiffer versions no additional iterations were necessary to converge to the grasping poses, the procedure also works for soft robots with stiffnesses in the range of $\sim 100\,\mathrm{N/rad} $.
In this case, the number of iterations increased by 30\%.
In all cases, the wall clock time for one loop of the optimizer increases by only one percent.
Because no additional calculation of the forward kinematics is necessary, and one pass through the torque loop is enough.

\section{Conclusions}
In this paper --  to our knowledge for the first time -- a calibration model including all DH parameters, joint and transversal elasticities was formulated and successfully calibrated for a humanoid robot.

For DLR's Agile Justin, the average error could be significantly reduced from prohibitively large 21\,mm to 3.1\,mm in the whole workspace. 
We provided a clear and concise derivation of the implicit kinematic model with elasticities from a torque equilibrium.
We showed that this complex implicit model can be used in optimization-based motion planning without any performance impact (for mildly elastic robots like Agile Justin). 
And even for a simulated soft robot with 20 times higher elasticities, the slowdown was only 30\%. 
We achieved this by tightly integrating the iterative solver for the implicit kinematic model into the optimization planner loop of the planner. 
Finally, we provided an in-depth discussion of the influence of the individual components of the model on the residual position error.

In the future, we want to make the calibration completely automatic and self-contained by replacing the use of an external tracking system by using the robot's head-mounted cameras, so combining this work with our previous work on automatic camera calibration~\cite{Birbach2015}.

\bibliographystyle{IEEEtranN-modified}
\bibliography{IEEEabrv, references.bib}

% Generated by IEEEtranN.bst, version: 1.14 (2015/08/26)
\begin{thebibliography}{17}
\providecommand{\natexlab}[1]{#1}
\providecommand{\url}[1]{#1}
\csname url@samestyle\endcsname
\providecommand{\newblock}{\relax}
\providecommand{\bibinfo}[2]{#2}
\providecommand{\BIBentrySTDinterwordspacing}{\spaceskip=0pt\relax}
\providecommand{\BIBentryALTinterwordstretchfactor}{4}
\providecommand{\BIBentryALTinterwordspacing}{\spaceskip=\fontdimen2\font plus
\BIBentryALTinterwordstretchfactor\fontdimen3\font minus
  \fontdimen4\font\relax}
\providecommand{\BIBforeignlanguage}[2]{{%
\expandafter\ifx\csname l@#1\endcsname\relax
\typeout{** WARNING: IEEEtranN.bst: No hyphenation pattern has been}%
\typeout{** loaded for the language `#1'. Using the pattern for}%
\typeout{** the default language instead.}%
\else
\language=\csname l@#1\endcsname
\fi
#2}}
\providecommand{\BIBdecl}{\relax}
\BIBdecl

\bibitem[Wagner et~al.(2013)Wagner, Frese, and Bauml]{Wagner2013}
R.~Wagner, U.~Frese, and B.~Bauml, ``{3D modeling, distance and gradient
  computation for motion planning: A direct GPGPU approach},'' in \emph{2013
  IEEE International Conference on Robotics and Automation}, no. Iii.\hskip 1em
  plus 0.5em minus 0.4em\relax IEEE, 5 2013, pp. 3586--3592.

\bibitem[Bauml et~al.(2014)Bauml, Hammer, Wagner, Birbach, Gumpert, Zhi,
  Hillenbrand, Beer, Friedl, and Butterfass]{Bauml2014}
B.~Bauml \emph{et~al.}, ``{Agile Justin: An upgraded member of DLR's family of
  lightweight and torque controlled humanoids},'' in \emph{2014 IEEE
  International Conference on Robotics and Automation (ICRA)}.\hskip 1em plus
  0.5em minus 0.4em\relax IEEE, 5 2014, pp. 2562--2563.

\bibitem[Roth et~al.(1987)Roth, Mooring, and Ravani]{Journal1987}
Z.~Roth, B.~Mooring, and B.~Ravani, ``{An overview of robot calibration},''
  \emph{IEEE Journal on Robotics and Automation}, vol.~3, no.~5, pp. 377--385,
  10 1987.

\bibitem[Pathre and Driels(1990)]{Pathre1990}
U.~S. Pathre and M.~R. Driels, ``{Simulation experiments in parameter
  identification for robot calibration},'' \emph{The International Journal of
  Advanced Manufacturing Technology}, vol.~5, no.~1, pp. 13--33, 2 1990.

\bibitem[Ginani and Motta(2011)]{Ginani2011}
L.~S. Ginani and J.~M. S.~T. Motta, ``{Theoretical and practical aspects of
  robot calibration with experimental verification},'' \emph{Journal of the
  Brazilian Society of Mechanical Sciences and Engineering}, vol.~33, no.~1,
  pp. 15--21, 3 2011.

\bibitem[Park and Kim(2011)]{Park2011}
I.~W. Park and J.~H. Kim, ``{Estimating entire geometric parameter errors of
  manipulator arm using laser module and stationary camera},'' \emph{IECON
  Proceedings (Industrial Electronics Conference)}, pp. 129--134, 2011.

\bibitem[Maier et~al.(2015)Maier, Wrobel, and Bennewitz]{Maier2015}
D.~Maier, S.~Wrobel, and M.~Bennewitz, ``{Whole-body self-calibration via
  graph-optimization and automatic configuration selection},'' in \emph{2015
  IEEE International Conference on Robotics and Automation (ICRA)}, vol.
  2015-June, no. June.\hskip 1em plus 0.5em minus 0.4em\relax IEEE, 5 2015, pp.
  5662--5668.

\bibitem[Stepanova et~al.(2019)Stepanova, Pajdla, and Hoffmann]{Stepanova2019}
K.~Stepanova, T.~Pajdla, and M.~Hoffmann, ``{Robot Self-Calibration Using
  Multiple Kinematic Chains-A Simulation Study on the iCub Humanoid Robot},''
  \emph{IEEE Robotics and Automation Letters}, vol.~4, no.~2, pp. 1900--1907,
  2019.

\bibitem[Klodmann et~al.(2011)Klodmann, Konietschke, Albu-Schaffer, and
  Hirzinger]{Klodmann2011}
J.~Klodmann, R.~Konietschke, A.~Albu-Schaffer, and G.~Hirzinger, ``{Static
  calibration of the DLR medical robot MIRO, a flexible lightweight robot with
  integrated torque sensors},'' in \emph{2011 IEEE/RSJ International Conference
  on Intelligent Robots and Systems}.\hskip 1em plus 0.5em minus 0.4em\relax
  IEEE, 9 2011, pp. 3708--3715.

\bibitem[Besset et~al.(2016)Besset, Olabi, and Gibaru]{Besset2016}
P.~Besset, A.~Olabi, and O.~Gibaru, ``{Advanced calibration applied to a
  collaborative robot},'' in \emph{2016 IEEE International Power Electronics
  and Motion Control Conference (PEMC)}.\hskip 1em plus 0.5em minus 0.4em\relax
  IEEE, 9 2016, pp. 662--667.

\bibitem[Caenen and Angue(1990)]{Caenen1990}
J.~Caenen and J.~Angue, ``{Identification of geometric and nongeometric
  parameters of robots},'' in \emph{Proceedings., IEEE International Conference
  on Robotics and Automation}.\hskip 1em plus 0.5em minus 0.4em\relax IEEE
  Comput. Soc. Press, 1990, pp. 1032--1037.

\bibitem[Khalil and Besnard(2002)]{Khalil2002}
W.~Khalil and S.~Besnard, ``{Geometric calibration of robots with flexible
  joints and links},'' \emph{Journal of Intelligent and Robotic Systems: Theory
  and Applications}, vol.~34, no.~4, pp. 357--379, 2002.

\bibitem[Lee(2013)]{Lee2013}
B.-J. Lee, ``{Geometrical Derivation of Differential Kinematics to Calibrate
  Model Parameters of Flexible Manipulator},'' \emph{International Journal of
  Advanced Robotic Systems}, vol.~10, no.~2, p. 106, 2 2013.

\bibitem[Zhou et~al.(2014)Zhou, Nguyen, and Kang]{Zhou2014b}
J.~Zhou, H.-N. Nguyen, and H.-J. Kang, ``{Simultaneous identification of joint
  compliance and kinematic parameters of industrial robots},''
  \emph{International Journal of Precision Engineering and Manufacturing},
  vol.~15, no.~11, pp. 2257--2264, 11 2014.

\bibitem[Carrillo et~al.(2013)Carrillo, Birbach, Taubig, Bauml, Frese, and
  Castellanos]{Carrillo2013}
H.~Carrillo \emph{et~al.}, ``{On task-oriented criteria for configurations
  selection in robot calibration},'' in \emph{2013 IEEE International
  Conference on Robotics and Automation}.\hskip 1em plus 0.5em minus
  0.4em\relax IEEE, 5 2013, pp. 3653--3659.

\bibitem[Birbach et~al.(2015)Birbach, Frese, and B{\"{a}}uml]{Birbach2015}
O.~Birbach, U.~Frese, and B.~B{\"{a}}uml, ``{Rapid calibration of a
  multi-sensorial humanoid's upper body: An automatic and self-contained
  approach},'' \emph{International Journal of Robotics Research}, vol.~34, no.
  4-5, pp. 420--436, 2015.

\bibitem[Birbach and Bauml(2014)]{Birbach2014}
O.~Birbach and B.~Bauml, ``{Calibrating a pair of inertial sensors at opposite
  ends of an imperfect kinematic chain},'' in \emph{2014 IEEE/RSJ International
  Conference on Intelligent Robots and Systems}, no. Iros.\hskip 1em plus 0.5em
  minus 0.4em\relax IEEE, 9 2014, pp. 422--428.

\end{thebibliography}

\end{document}